\documentclass[sigconf,prepirnt]{acmart}

\AtBeginDocument{%
  }

\usepackage[table]{xcolor}
\usepackage{times}
\usepackage{latexsym}
\usepackage{url}
\usepackage{graphicx} 
\usepackage{url}
\usepackage{multicol}
\usepackage{multirow}
\usepackage[ruled,vlined]{algorithm2e}
\usepackage{enumitem} 
\usepackage{subcaption}

\usepackage[most]{tcolorbox} 
\tcbuselibrary{listingsutf8}

\usepackage{fvextra}  
\newtcolorbox{textbox}{
  enhanced,
  colback=gray!5,
  colframe=gray!50,
  boxrule=0.5pt,
  arc=4pt,
  auto outer arc,
  fontupper=\scriptsize
}

\usepackage{booktabs}
\usepackage{xcolor}
\usepackage{bbding}
\usepackage{graphicx} 

\definecolor{mygreen}{RGB}{34,188,34}
\definecolor{myred}{RGB}{222,34,34}
\newcommand{\cmark}{\textcolor{mygreen}{\CheckmarkBold}}
\newcommand{\xmark}{\textcolor{myred}{\XSolidBrush}}
\usepackage{xspace}
\usepackage{gradient-text}
\newcommand{\finauditing}{\gradientRGB{FinAuditing}{20,20,185}{10,180,85}\xspace}

\begin{document}

\title{\textsc{\finauditing}: A Financial Taxonomy-Structured Multi-Document Benchmark for Evaluating LLMs}

\author{Yan Wang}
\affiliation{%
  \institution{\normalsize The Fin AI}
  \country{\normalsize USA}
}
\email{wy2266336@gmail.com}

\author{Keyi Wang}
\affiliation{%
  \institution{\normalsize Columbia University}
  \country{\normalsize USA}}

\author{Shanshan Yang}
\affiliation{%
  \institution{\normalsize Stevens Institute of Technology }
  \country{\normalsize USA}
}

\author{Jaisal Patel}
\affiliation{%
 \institution{\normalsize Rensselaer Polytechnic Institute}
 \country{\normalsize USA}}

\author{Jeff Zhao}
\affiliation{%
  \institution{\normalsize UT Austin}
  \country{\normalsize USA}}

\author{Fengran Mo}
\affiliation{%
  \institution{\normalsize University of Montreal}
  \country{\normalsize Canada}}

\author{Xueqing Peng}
\authornote{Corresponding authors.}
\author{Lingfei Qian}
\authornotemark[1]
\affiliation{%
  \institution{\normalsize The Fin AI}
  \country{\normalsize USA}
}

\author{Yankai Chen}
\authornotemark[1]
\affiliation{%
  \institution{\normalsize McGill University \\ MBZUAI}
  \country{\normalsize Canada}}
\email{yankaichan3@gmail.com}

\author{V\'ictor Guti\'errez-Basulto}
\authornotemark[1]
\affiliation{%
  \institution{\normalsize Cardiff University}
  \country{\normalsize UK}}
\email{GutierrezBasultoV@cardiff.ac.uk}

\author{Jimin Huang}
\affiliation{%
   \institution{The University of Manchester}
   \city{Manchester}
   \country{United Kingdom}
}
\affiliation{%
  \institution{\normalsize The Fin AI}
  \country{\normalsize USA}
}

\author{Guojun Xiong}
\affiliation{%
  \institution{\normalsize Harvard University}
  \country{\normalsize USA}}

\author{Xiao-Yang Liu}
\affiliation{%
  \institution{\normalsize Columbia University}
  \country{\normalsize USA}}

\author{Xue (Steve) Liu}
\affiliation{%
  \institution{\normalsize McGill University \\ MBZUAI}
  \country{\normalsize Canada}}

\author{Jian-Yun Nie}
\affiliation{%
  \institution{\normalsize University of Montreal}
  \country{\normalsize Canada}}

\renewcommand{\shortauthors}{Yan et al.}

\begin{abstract}
Going beyond simple text processing, financial auditing requires detecting semantic, structural, and numerical inconsistencies across large-scale disclosures. As financial reports are filed in XBRL, a structured XML format governed by accounting standards, auditing becomes a structured information extraction and reasoning problem involving concept alignment, taxonomy-defined relations, and cross-document consistency. Although large language models (LLMs) show promise on isolated financial tasks, their capability in professional-grade auditing remains unclear. We introduce \textsc{\finauditing}, a taxonomy-aligned, structure-aware benchmark built from real XBRL filings. It contains 1,102 annotated instances averaging over 33k tokens and defines three tasks: \emph{Financial Semantic Matching} (FinSM), \emph{Financial Relationship Extraction} (FinRE), and \emph{Financial Mathematical Reasoning} (FinMR). Evaluations of 13 state-of-the-art LLMs reveal substantial gaps in concept retrieval, taxonomy-aware relation modeling, and consistent cross-document reasoning. These findings highlight the need for realistic, structure-aware benchmarks. We release the evaluation code\footnote{\url{https://github.com/The-FinAI/FinAuditing}} and dataset\footnote{\url{https://huggingface.co/collections/TheFinAI/finauditing}} publicly, and the task currently serves as the official benchmark of an ongoing public evaluation contest\footnote{\url{https://open-finance-lab.github.io/SecureFinAI_Contest_2026/}}.
\end{abstract}




\maketitle

\begin{figure}[!htbp]
  \centering
  \setlength{\abovecaptionskip}{0pt}
  \setlength{\belowcaptionskip}{0pt}
  \includegraphics[width=\linewidth]{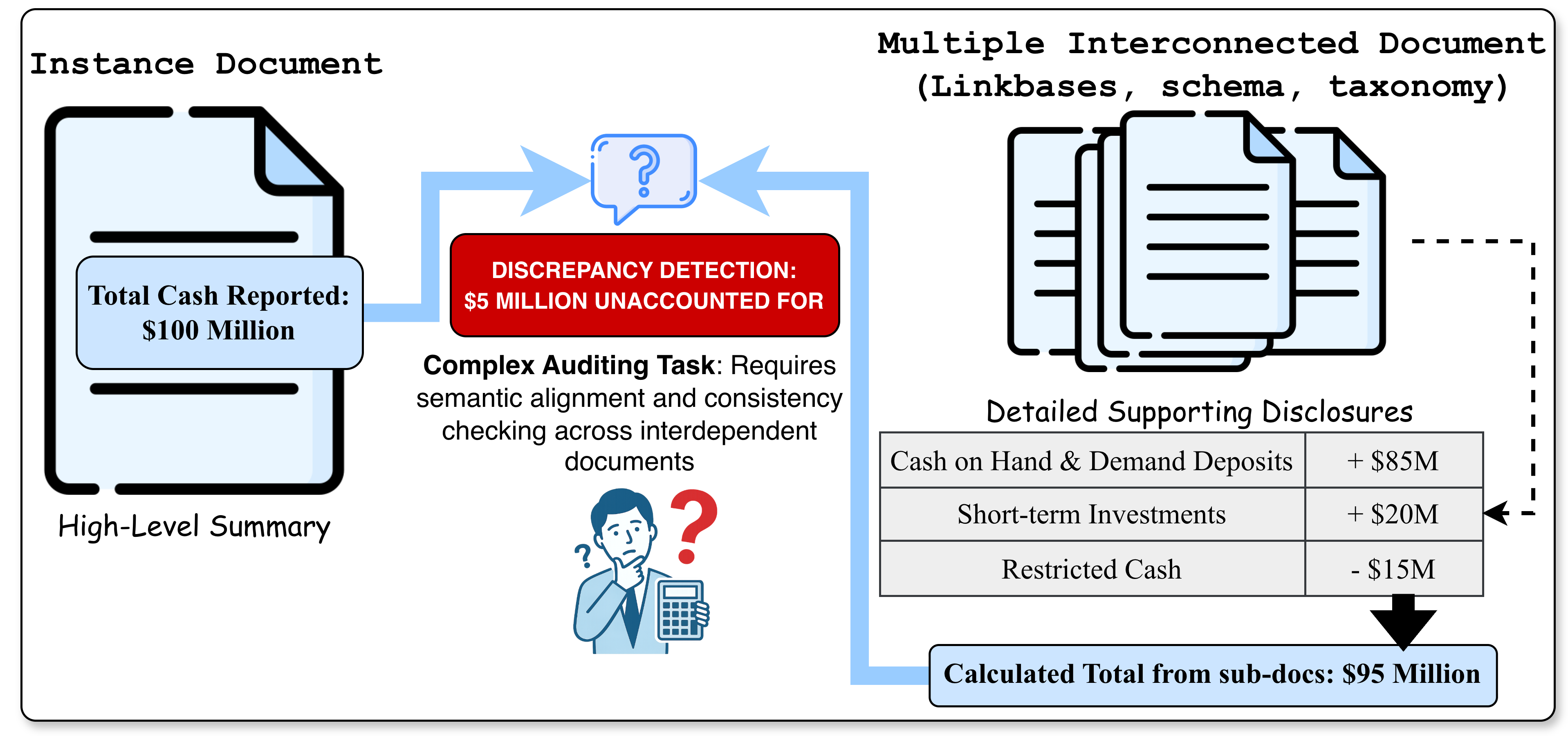}
  \caption{An illustrative example of cross-document financial inconsistency.}
  \label{fig:err}
\end{figure}

\section{Introduction}

Consider a scenario where a global corporation reports ``Total Cash'' as \$100M in its high-level summary, while detailed supporting disclosures across sub-documents sum to only \$95M. 
Detecting such inconsistencies is a non-trivial information verification problem.
In modern financial reporting, this information spans thousands of pages of structured filings, making manual verification nearly impossible. 
While ensuring data accuracy is vital for market transparency, identifying these ``hidden'' inconsistencies remains challenging even for advanced AI systems, due to the massive scale and structural complexity involved.

Modern financial disclosures are governed by Generally Accepted Accounting Principles (GAAP) and operationalized via the eXtensible Business Reporting Language (XBRL) \cite{debreceny2010does}. 
Unlike plain text, these filings are heterogeneous, hierarchical, and highly interdependent \cite{debreceny2010does,hoitash2018measuring}. 
Auditing these documents requires more than keyword search, as it demands sophisticated structured information retrieval (SIR) and reasoning~\cite{li2015literature,da2018xaudit}. 
Models must semantically align concepts across disparate tables, interpret nested hierarchical metadata, and maintain logical consistency across multiple interdependent documents. When these capabilities fall short, the consequences are tangible: firms issue restatements, i.e., costly corrections of previous reports, which have risen persistently as disclosure complexity outpaces current analytical tools\footnote{\url{https://www.ft.com/content/716c4ad5-e8fa-4a34-afba-9fb2d1db019d}}.

\begin{table}[!t]
\centering
\caption{Comparison of existing financial NLP benchmarks and \textsc{\finauditing}. We compare support for structured retrieval (Struct.Retrieval), hierarchical information extraction (Hiera.IE), multi-step numerical reasoning (MultiStep.Reasoning), taxonomy-driven supervision (Need.Taxonomy), and cross-document reasoning (At.Cross-Doc).}
\label{tab:benchmark_comparison}
\resizebox{\linewidth}{!}{%
\begin{tabular}{l|c|c|c|c|c}
\hline
\textbf{Benchmark} & \textbf{Struct.Retr.} & \textbf{Hiera.IE} & \textbf{MultiStep.Reas.} & \textbf{Need.Taxon.} & \textbf{At.Cross-Doc} \\ \hline
FiNER \cite{loukas2022finer} & \xmark & \xmark & \xmark & \xmark & \xmark \\
FNXL \cite{sharma2023financial} & \xmark & \xmark & \xmark & \xmark & \xmark \\
FinTagging \cite{wang2025fintagging} & \cmark & \xmark & \xmark & \cmark & \xmark \\
FinQA \cite{chen2021finqa} & \xmark & \xmark & \cmark & \xmark & \xmark \\
ConvFinQA \cite{chen2022convfinqa} & \xmark & \xmark & \cmark & \xmark & \xmark \\
TAT-QA \cite{zhu2021tat} & \xmark & \xmark & \cmark & \xmark & \xmark \\
MultiHiertt \cite{zhao2022multihiertt} & \xmark & \xmark & \cmark & \xmark & \xmark \\
DOCMATH-EVAL \cite{zhao2023docmath} & \xmark & \xmark & \cmark & \xmark & \xmark \\
\rowcolor[gray]{.9} \textbf{FinAuditing (ours)} & \cmark & \cmark & \cmark & \cmark & \cmark \\ \hline
\end{tabular}%
}
\end{table}

Recent advances in large language models (LLMs) have sparked interest in applying them to financial tasks, such as information extraction \cite{loukas2022finer,sharma2023financial,wang2025fintagging}, question answering \cite{chen2021finqa,chen2022convfinqa,zhu2021tat,zhao2022multihiertt,zhao2023docmath}, and numerical reasoning \cite{soun2022accurate} and trading \cite{qian2025agents,li2025investorbench}.
However, it remains unclear whether these models can support \emph{professional-grade auditing}, which requires consistent reasoning over hierarchical structures, taxonomy-defined relations, and cross-document dependencies.
This raises a critical question: \emph{how should we evaluate models on structured, regulation-driven financial information tasks?}
As summarized in Table~\ref{tab:benchmark_comparison}, existing financial NLP benchmarks exhibit three fundamental limitations:
(1) \textbf{Contextual Isolation}: Most benchmarks operate on isolated text snippets or tables, failing to capture the cross-document dependencies that are central to real financial filings \cite{zhao2022multihiertt}. 
(2) \textbf{Surface-Level Evaluation}: Prior work predominantly evaluates local semantic or numerical correctness, without assessing multi-step structural reasoning over reporting hierarchies \cite{zhao2023docmath}. 
(3) \textbf{Taxonomy Underutilization}: Although official GAAP taxonomies define valid financial concepts and relations, few works evaluate whether models can retrieve and reason over these  constraints \cite{wang2025fintagging}.



To address these gaps, we introduce \textsc{\finauditing}, a taxonomy-aligned, structure-aware, multi-document benchmark derived from real-world US-GAAP-compliant XBRL filings.
\textsc{\finauditing} reframes financial auditing as a structured information reasoning problem, and defines three complementary subtasks:
\textbf{Financial Semantic Matching (FinSM)} for concept-level semantic alignment,
\textbf{Financial Relationship Extraction (FinRE)} for structural and relational consistency,
and \textbf{Financial Mathematical Reasoning (FinMR)} for taxonomy-grounded numerical verification.
Together, these tasks shift evaluation from isolated text understanding to holistic reasoning over structured financial information, providing a high-fidelity testbed for financial capabilities under formal constraints.
Our main contributions are threefold:
\begin{itemize}[leftmargin=*]
\item \textbf{A Taxonomy-Grounded Financial Resource}: We release the first benchmark built from real-world XBRL filings that explicitly integrates multi-document structures and official GAAP taxonomies for evaluation.

\item \textbf{Structure-Aware Task Design}: We propose three complementary subtasks (\textit{FinSM, FinRE, FinMR}) that capture semantic, relational, and numerical reasoning challenges inherent to structured financial information.

\item \textbf{Systematic Evaluation of Intelligent Systems}: We provide an initial evaluation of 13 SOTA LLMs, while emphasizing that \textsc{\finauditing} is model-agnostic and designed to support the evaluation of complex architectures, including retrieval-augmented generation (RAG) pipelines and multi-agent auditing systems, for structured financial information retrieval and reasoning. 
\end{itemize}

\section{Related work}
Prior financial NLP research has advanced information extraction~\cite{loukas2022finer,sharma2023financial,wang2025fintagging}, semantic understanding~\cite{mukherjee2022ectsum,xie2023pixiu,sinha2021impact}, and numerical reasoning~\cite{chen2021finqa,chen2022convfinqa,zhu2021tat,zhao2022multihiertt,zhao2023docmath}. However, most benchmarks focus on unstructured or semi-structured data from individual documents, such as financial reports or news articles, and overlook the structured, hierarchical, and interconnected nature of XBRL filings. Early XBRL-focused studies primarily address localized subtasks, including numerical entity extraction~\cite{loukas2022finer}, sentence-level numeral labeling~\cite{sharma2023financial}, tag normalization~\cite{wang2025fintagging}, and retrieval- or agent-based evaluation of LLMs on XBRL data~\cite{katz2024information,han2024xbrl}. While informative, these efforts evaluate narrow components in isolation and do not capture the taxonomy-level dependencies that characterize real-world XBRL structures. Related benchmarks on tabular and textual reasoning, such as FinQA~\cite{chen2021finqa}, ConvFinQA~\cite{chen2022convfinqa}, TAT-QA~\cite{zhu2021tat}, MultiHiertt~\cite{zhao2022multihiertt}, and DocMath-Eval~\cite{zhao2023docmath}, similarly lack taxonomy-driven supervision. In contrast, \textsc{\finauditing} is designed to evaluate LLMs on structured, hierarchical, and cross-document reasoning grounded in realistic XBRLs.

\begin{figure}[!h]
    \centering
    \setlength{\abovecaptionskip}{2pt}
    \includegraphics[width=.9\linewidth]{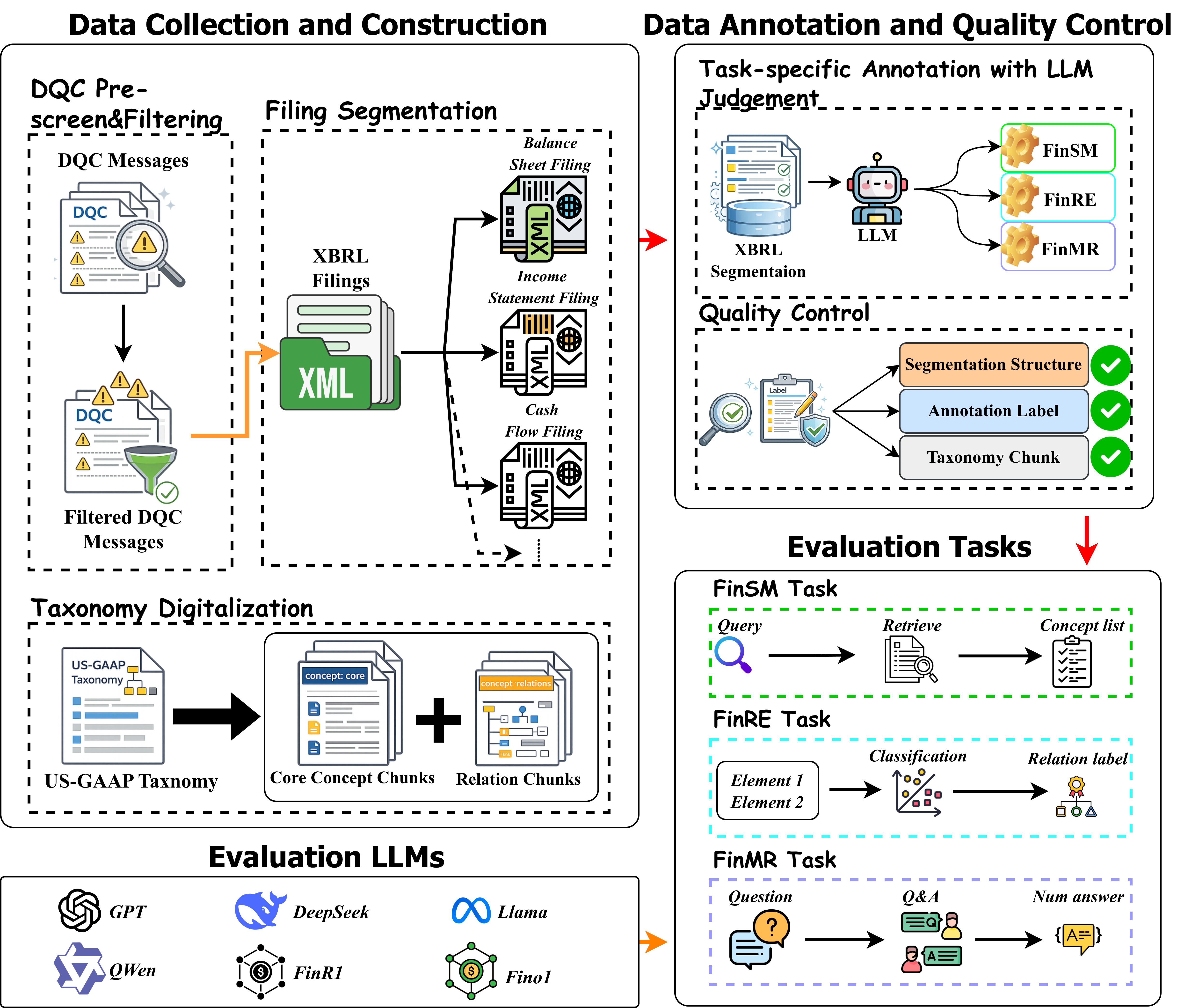}
    \caption{Overview of the \textsc{\finauditing} construction pipeline.}
    \label{fig:framework}
\end{figure}

\section{\finauditing}
Detecting errors in structured financial documents requires semantic, structural, and numerical reasoning. 
As shown in Figure~\ref{fig:framework}, we introduce \textsc{\finauditing}, a benchmark for evaluating LLMs in identifying inconsistencies across multi-document XBRL filings.

\subsection{Task Formulation}
\label{task_form}
Given a target XBRL filing consisting of six interconnected, XML documents $\mathcal{D}=\{d_1, d_2, d_3, d_4, d_5, d_6\}$ (described in Table~\ref{tab:component}) and a reference US-GAAP taxonomy $\mathcal{T}$, the task is to identify errors $\mathcal{E}$ within the filing. \textbf{These errors include semantic-matching inconsistencies, relationship-extraction failures, and mathematical-reasoning violations}. 
Models must analyze the filing's internal structure and semantics while leveraging domain knowledge from the reference taxonomy. We formulate this as follows:
\begin{equation}
    f: \left( \mathcal{D}, \mathcal{T} \right) \mapsto \mathcal{E}
\end{equation}
where $\mathcal{E}$ is the specific description of the errors found in the $\mathcal{D}$.
 
To operationalize error detection across XBRL filings, we identify three fundamental capabilities: \textbf{semantic consistency}, \textbf{structural relationship understanding}, and \textbf{numerical reasoning}. 
Accordingly, \textsc{\finauditing} defines three tasks: (1) \textbf{Financial Semantic Matching (FinSM)} identifies mismatches between textual mentions in filings and standardized taxonomy labels; 
(2) \textbf{Financial Relationship Extraction (FinRE)} detects errors in hierarchical and compositional relationships among financial items; and (3) \textbf{Financial Mathematical Reasoning (FinMR)} verifies numerical correctness based on accounting logic and contextual inference.

\subsubsection{{FinSM}}
The semantic matching task evaluates whether models can identify financial tags in XBRL filings that misalign with standardized US-GAAP taxonomy concepts.  
Given a filing with multiple candidate tag assignments, the objective is to detect semantically inconsistent or inappropriate concept assignments within a set of complex, structured documents.

\begin{tcolorbox}[
    colback=lightgray!10,
    colframe=black,
    title=FinSM Example,
    title style={fontupper=\scriptsize},
    width=\linewidth,       
    boxrule=0.4pt,
    arc=2pt,
    top=1pt,
    bottom=1pt,
    left=1pt,
    right=1pt,
    boxsep=1.5pt,
    before skip=4pt, 
    after skip=4pt
]
\scriptsize
\begin{lstlisting}[breaklines=true, basicstyle=\ttfamily\scriptsize]
You are an auditor for XBRL filings. Given the question and the provided filing (schema, presentation, calculation, definition, label, instance, and US-GAAP taxonomy), identify erroneous US-GAAP concepts.
You must reference the US-GAAP taxonomy to retrieve and check elements across the entire filing. 
Output only a JSON array of strings, each string is one erroneous concept.
Output Example:
["us-gaap:Revenues", "us-gaap:OperatingIncomeLoss"]
No explanations. No extra text. Only the JSON array.
\end{lstlisting}
\end{tcolorbox}

This task follows an information retrieval paradigm. Given
(1) a query $Q$ describing a financial term related to currency or credit risk concentration, (2) an XBRL filing $\mathcal{D}$, and (3) the US-GAAP taxonomy $\mathcal{T}$. The goal is to retrieve the set of mismatched US-GAAP tags within the filing, denoted as $\mathcal{E}$.
For the \textbf{FinSM} task, we evaluate retrieval performance using metrics: 
\textit{Hit Rate} ($HR@k$)~\cite{he2017neural},
\textit{Recall} ($R@k$), and
\textit{Macro-F1} ($MF1@k$),
with $k \in \{1,5,10,20\}$.



\subsubsection{{FinRE}}

The relationship extraction task evaluates whether models can identify structural errors among financial elements by interpreting hierarchical and compositional dependencies between the filing and the external taxonomy. 
This task assesses the LLMs' capacity for semantic understanding and structured data interpretation within the financial reporting context.

\begin{tcolorbox}[
    colback=lightgray!10,
    colframe=black,
    title=FinRE Example,
    title style={fontupper=\scriptsize},
    width=\linewidth,
    boxrule=0.3pt,        
    arc=1.5pt,            
    top=0.5pt,            
    bottom=0.5pt,
    left=0.5pt,
    right=0.5pt,
    boxsep=1pt,
    before skip=1pt, 
    after skip=1pt
]
\scriptsize               
\begin{lstlisting}[breaklines=true, basicstyle=\ttfamily\scriptsize]
You are an auditor for XBRL filings. Given two concepts and the provided filing (schema, presentation, calculation, definition, label, instance, and US-GAAP taxonomy), determine which one of the following erroneous relationship types best describes the relationship between the two concepts:
(1) Reversal Definition: <definition>.
(2) Inappropriateness Definition: <definition>.
(3) CombinationErr Definition: <definition>.
You must reference the US-GAAP taxonomy to retrieve and check the two concepts across the entire filing.
Output only one of the following labels exactly, chosen strictly from:["Reversal", "Inappropriateness", "CombinationErr"].
No explanations. No extra text. Only the label.
\end{lstlisting}
\end{tcolorbox}

Formally, given a relation error type space $L$ with predefined semantic definitions,
an XBRL filing $\mathcal{D}$, and the US-GAAP taxonomy $\mathcal{T}$,
the task aims to classify the specific relation error type $\ell \in L$ between two specific elements $e_1$ and $e_2$, which are derived from the current XBRL filing $\mathcal{D}$,
based on the information provided by $\mathcal{T}$.

We define three relation error types $\{\ell_1,\ell_2,\ell_3\}$ in $L$:
(1) \textit{Reversal}, which occurs when the hierarchical relationship between parent and child elements is mistakenly reversed;
(2) \textit{Inappropriateness}, which indicates that a child element is incorrectly associated with an inappropriate parent element; and
(3) \textit{CombinationErr}, which refers to an invalid combination of axis and member elements that violates the structural constraints defined in the taxonomy.
We use the metrics, including
\textit{Accuracy} (Acc),
\textit{Macro Precision} (P),
\textit{Macro Recall} (R), and
\textit{Macro F1}, to evaluate the performance of this task.

\subsubsection{{FinMR}}
The mathematical reasoning task focuses on inferring correct numerical values among financial elements based on structured representations, calculation hierarchies, and constraints defined in the XBRL filing. The objective is to assess whether reported values are internally consistent and compliant with domain-specific standards.

\begin{tcolorbox}[
    colback=lightgray!10,
    colframe=black,
    title=FinMR Example,
    title style={fontupper=\scriptsize},
    width=\linewidth,       
    boxrule=0.4pt,
    arc=2pt,
    top=1pt,
    bottom=1pt,
    left=1pt,
    right=1pt,
    boxsep=1.5pt,              
    before skip=4pt, 
    after skip=4pt
]
\scriptsize
\begin{lstlisting}[breaklines=true, basicstyle=\ttfamily\scriptsize]
You are an auditor for XBRL filings. Given the question and the provided filing (schema, presentation, calculation, definition, label, instance, and US-GAAP taxonomy), identify the reported value of a financial element and calculate the actual value that should be reported based on calculation relationships.
Answer strictly in the following JSON format:
{"extracted_value": "<numeric value reported in the instance document, or 0 if not found, keep the same format as in the XBRL filing>",
  "calculated_value": "<numeric value computed from calculation relationships, or 0 if not computable, use the same number formatting style as extracted_value>"}
No explanations. No extra text. Only the JSON object.
\end{lstlisting}
\end{tcolorbox}

Specifically, the task involves an XBRL filing $\mathcal{D}$, the US-GAAP taxonomy $\mathcal{T}$, and a pair of questions ($q_1$, $q_2$). 
Here, $q_1$ requests the extraction of a reported value and
$q_2$ pertains to the computation of the corresponding true value.
The model is required to: (1) extract the reported numeric value $v$ for a given instance from $\mathcal{D}$, and
(2) compute the corresponding real value $\mu$ based on the structural and numerical information in $\mathcal{D}$ and $\mathcal{T}$.
The extracted value $v$ and the computed value $\mu$ are then compared to determine whether the reported value is correct.
The output of this task is a structured answer containing both $v$ and $\mu$, represented in a JSON format.
For this task, we evaluate the overall \textit{Accuracy} using an LLM-as-a-judge framework~\cite{gu2024survey,li2024llms}, together with three complementary error indicators.
The structural error rate (\textit{SER}) measures cases where the model fails to produce a valid output structure in the required JSON format.
The extraction error rate (\textit{EER}) captures errors in identifying the correct numerical value to be extracted, even when the output structure is valid.
The calculation error rate (\textit{CER}) reflects mistakes in numerical computation, where the final calculated value is incorrect despite the correct structure and extraction.

\subsection{Data Collection and Construction}

\textsc{\finauditing} is established as a high-fidelity testbed to catalyze research on \textbf{taxonomy-grounded reasoning} and \textbf{structured retrieval} within complex, multi-document financial environments. 
As illustrated in Figure~\ref{fig:framework}, our resource bridges the gap between raw XML data and the semantic constraints of the US-GAAP taxonomy.

\subsubsection{Authoritative Error Signals from DQC Messages}
\label{dqc_message}
To identify reliable auditing errors, we ground our benchmark construction in authoritative error signals released by the Data Quality Committee (DQC)\footnote{\url{https://xbrl.us/data-quality/filing-results/}}. 
Our construction follows a two-step pipeline:
(1) \textbf{Pre-screening}: We collected 4,545 official DQC error messages for 372 companies between 2020 and 2024 via the XBRL US portal. 
Each message links to a deterministic DQC US rule, providing unambiguous labels for semantic, relational, or numerical inconsistencies.
(2) \textbf{Filtering}: We selected the 9 most frequent DQC error types, covering 60.33\% of all cases (Table~\ref{tab:statistic_err}), and retrieved 218 corresponding XBRL filings from the SEC EDGAR database, the official repository for public company disclosures.

\begin{table}[!t]
\setlength{\abovecaptionskip}{3pt}  
 \caption{Distribution of common error types in XBRL filings.}
 \label{tab:statistic_err}
 \centering
 \resizebox{\linewidth}{!}{%
 \begin{tabular}{llccc}
 \hline
 \textbf{Broad Category} & \textbf{Error Type} & \textbf{\# Cases} & \textbf{Proportion (\%)} & \textbf{DQC US ID} \\ \hline
 Semantic-based & FS with no associated calculation & 386 & 8.49 & DQC\_0099 \\
                & Concentration risk & 217 & 4.77 & DQC\_0109 \\
                & Location axis with a single member & 208 & 4.58 & DQC\_0137 \\ \hline
 Relation-based & Sibling and child relationships & 570 & 12.54 & DQC\_0081 \\
                & Axis with inappropriate members & 313 & 6.89 & DQC\_0001 \\
                & Inappropriate cash flow presentation & 219 & 4.82 & DQC\_0145 \\ \hline
 Calculation-based & FS calculation check with no dimensional data & 260 & 5.72 & DQC\_0126 \\
                   & Negative values & 368 & 8.10 & DQC\_0015 \\
                   & FS tables dimensional cross check & 202 & 4.44 & DQC\_0117 \\ \hline
 \multicolumn{2}{c}{\textbf{Total}} & \textbf{2,743} & \textbf{60.33} & - \\ \hline
 \end{tabular}%
 }
\end{table}

\subsubsection{XBRL Filing Segmentation}
\label{xbrl_seg}
Using ``SEC-url'' identifiers from DQC messages (Section~\ref{dqc_message}), we retrieve the corresponding XBRL filings as the primary data representation for structured financial reporting. 
An XBRL filing comprises multiple interconnected XML documents that jointly encode both financial facts and their structural constraints. 
As summarized in Table~\ref{tab:component}, each filing contains: an \textit{instance document} with reported values, a \textit{schema document} defining element types, and several \textit{linkbases} (calculation, presentation, definition, and label) that specify hierarchical organization, arithmetic relationships, and semantic metadata \cite{debreceny2010does, hoitash2018measuring}.

Due to this rich but highly interdependent structure, complete XBRL filings are often extremely long and difficult to process directly, posing challenges for both manual inspection and long-context LLM evaluation.
To balance document length with structural fidelity, we perform statement-level filing segmentation. 
Specifically, we extract \texttt{roleType} identifiers from the schema document to identify distinct financial statements (e.g., balance sheets or income statements). 
We then leverage the \textbf{presentation linkbase} to map the hierarchical layout of each statement to its corresponding facts in the instance document. 
This segmentation preserves the original XML dependencies across the schema, linkbases, and instance documents, while producing manageable, structure-aware sub-filings suitable for downstream auditing tasks and long-context LLM processing.

\begin{table}[t]
\setlength{\abovecaptionskip}{3pt}  
\caption{Main Components of XBRL Filings}
\label{tab:component}
\small
\centering
\resizebox{0.9\columnwidth}{!}{%
\begin{tabular}{l|p{6.5cm}}
\hline
\textbf{Component} & \textbf{Description} \\
\hline
Instance Document & Actual financial data structured according to a predefined taxonomy. \\
\hline
Schema Document & Defines the structure, classification, and data types of financial elements. \\
\hline
Calculation Linkbase & Specifies arithmetic relationships between elements to ensure logical consistency. \\
\hline
Presentation Linkbase & Organizes financial data into a human-readable structure for reporting. \\
\hline
Definition Linkbase & Captures semantic relationships between elements, such as component relations. \\
\hline
Label Linkbase & Provides human-readable names and descriptions, supporting multiple languages. \\
\hline
\end{tabular}%
}
\end{table}

\subsubsection{US-GAAP Taxonomy Digitalization}
\label{taxonomy_chunking}
To facilitate structured access to the US-GAAP Taxonomy, we digitalize it into concept-centric chunks. 
We extract each concept and its associated semantic information from the official taxonomy files (e.g., GAAP Taxonomy 2024.xlsx), parsing all relevant sheets: Concepts, Presentation, Calculation, Definition, Reference, and Enumeration.
We organized the extracted information into two major chunk types: \textbf{core chunks} contain each concept’s intrinsic attributes, such as label, type, balance, period type, and documentation.
\textbf{relation chunks} encode structural and semantic links to other concepts, including hierarchical presentation paths, parent–child calculation relationships, definitional arcs, and external references. 
This digitalization process preserves the taxonomy's hierarchical and relational structure while enabling fine-grained concept-level indexing, retrieval, and reasoning.

\subsection{Data Annotation and Quality Control}

\textsc{\finauditing} provides 1,102 annotated instances, each averaging over 33k tokens (Table~\ref{tab:task-stats}), capturing the long-context characteristics of practical financial auditing scenarios.

\begin{table}[!t]
\setlength{\abovecaptionskip}{3pt}  
 \caption{Resource statistics across tasks (tokens via cl100k\_base).}
 \label{tab:task-stats}
 \centering
 \tiny
 \resizebox{.85\linewidth}{!}{%
 \begin{tabular}{l|ccccc}
 \hline
 Task & Total Tokens & Max Tokens & Avg. Tokens & Std. Tokens & Instances \\ \hline
 FinSM & 11,170,024 & 62,777 & 33,848.56 & 11,402.49 & 330 \\
 FinRE & 14,819,627 & 65,030 & 33,680.97 & 10,847.23 & 440 \\
 FinMR & 11,845,255 & 59,641 & 35,678.48 & 9,852.73 & 332 \\
 \hline
 Total & 37,834,906 & 65,030 & 34,338.64 & 10,700.82 & 1,102 \\
 \hline
 \end{tabular}%
 }
\end{table}


\subsubsection{Annotation via LLM-Assisted Parsing} 
Ground-truth supervision is deterministically derived from DQC error messages. 
We employ GPT-4o-mini solely as a high-efficiency parser, rather than a decision-maker, to convert these messages into structured annotations. 
The model extracts (i) DQC-defined ground-truth labels, (ii) task-specific elements for formulating queries, relations, or questions, and (iii) evidence pointers linking errors to relevant XBRL sub-filings and US-GAAP taxonomy chunks. 
These pointers enable the retrieval of the corresponding XBRL segments and taxonomy chunks, grounding each instance in both the reported filing and the governing rules. 
Task-specific instances are then constructed as follows:
(1) \textbf{FinSM}: Semantically incorrect US-GAAP concepts identified in DQC messages serve as ground-truth answers. 
The corresponding statement-level XBRL sub-filings are retrieved as input.
(2) \textbf{FinRE}: DQC rule IDs are deterministically mapped to relation labels (DQC\_0081: \textit{Reversal}, DQC\_0001: \textit{CombinationErr}, DQC\_0145: \textit{Inappropriateness}). 
Participating elements form $(\textit{head}, \textit{relation}, \textit{tail})$ triples, where the relation is the ground-truth label and the associated XBRL sub-filings provide input context.
(3) \textbf{FinMR}: 
DQC messages contain both reported values and correct values implied by calculation linkbases.
We use the target concept and reporting period to construct the input questions, with statement-level XBRL sub-filings serving as the input for numerical reasoning.

\subsubsection{Quality Control} 
To ensure the reliability and consistency of the \textsc{\finauditing} benchmark, we conducted a multi-stage quality control process: 
(1) \textbf{Structural Verification}: We manually inspected 50 randomly sampled filings to verify that extracted statement segments correctly matched their \texttt{roleType} identifiers and preserved the original reporting hierarchy. 
(2) \textbf{Annotation Validation}: 50\% of all instances underwent double-validation, where GPT-4o-mini outputs were compared against independent human review, with all discrepancies corrected to ensure ground-truth accuracy. 
(3) \textbf{Referential Integrity}: We verified that all relation chunks reference valid core concepts in the taxonomy, ensuring the integrity of taxonomy-grounded reasoning paths.

\subsection{Evaluated LLMs}
Our goal is to evaluate the foundational capabilities of state-of-the-art LLMs on the \textsc{\finauditing} benchmark, to assess their capabilities and limitations in financial auditing.
To this end, we select a diverse set of models, including one closed-source general-purpose LLM (GPT-4o~\cite{hurst2024gpt}), ten open-source general-purpose LLMs of varying sizes (DeepSeek-V3~\cite{liu2024deepseek}, Qwen3-235B-A22B-Instruct-2507~\cite{qwen3technicalreport}, Llama-4-Scout-17B-16E-Instruct~\cite{meta2025llama}, Qwen2.5-72B-Instruct~\cite{qwen2}, Llama-3.3-70B-Instruct~\cite{grattafiori2024llama}, Qwen3-32B~\cite{qwen3technicalreport}, Gemma-3-27B-IT~\cite{gemma_2024}, Gemma-3-12B-IT~\cite{gemma_2024}, Llama-3.1-8B-Instruct~\cite{grattafiori2024llama}, and Llama-3.2-3B-Instruct~\cite{grattafiori2024llama}),
as well as two open-source financial LLMs (Fino1~\cite{qian2025fino1} and FinR1~\cite{liu2025finr1largelanguagemodel}).
All evaluations use the LM Evaluation Harness~\cite{eval-harness}.
Proprietary models are accessed via APIs, while open-source models are run locally on a 4$\times$H200 GPU cluster.
For all tasks, we standardize the maximum input length to 81,920 tokens and the generation limit to 512 tokens.

\section{Benchmarking Results and Empirical Analysis}
\label{sec:res}

\subsection{FinSM}


\begin{table}[!t]
\setlength{\abovecaptionskip}{3pt}
\caption{Zero-shot performance (\%) on the FinSM task.}
\label{tab:sm-all}
\centering
\resizebox{.9\linewidth}{!}{%
\begin{tabular}{l|ccc|ccc|ccc}
\hline
 & \multicolumn{3}{c|}{Hit Rate} 
 & \multicolumn{3}{c|}{Recall} 
 & \multicolumn{3}{c}{Macro-F1} \\
LLMs 
& $@5$ & $@10$ & $@20$
& $@5$ & $@10$ & $@20$
& $@5$ & $@10$ & $@20$ \\
\hline
GPT-4o 
& 9.09 & 9.09 & 9.09
& 6.82 & 6.98 & 7.01
& 6.91 & 7.03 & 7.02 \\
DeepSeek-V3 
& 11.82 & 11.82 & 12.42
& 8.82 & 9.33 & 10.11
& 9.54 & 9.98 & 10.17 \\
Qwen3-235B-A22B-Instruct-2507 
& 10.00 & 10.00 & 10.00
& 7.77 & 8.17 & 8.33
& 7.83 & 7.96 & 7.97 \\
Llama-4-Scout-17B-16E-Instruct 
& 2.42 & 2.73 & 3.03
& 1.58 & 1.91 & 2.21
& 1.55 & 1.76 & 1.80 \\
Qwen2.5-72B-Instruct 
& 8.18 & 8.48 & 8.48
& 4.81 & 5.18 & 5.18
& 5.29 & 5.42 & 5.42 \\
Llama-3.3-70B-Instruct 
& 5.15 & 5.15 & 5.15
& 3.78 & 3.85 & 3.85
& 4.07 & 4.14 & 4.14 \\
Qwen3-32B 
& 0.00 & 0.00 & 0.00
& 0.00 & 0.00 & 0.00
& 0.00 & 0.00 & 0.00 \\
gemma-3-27b-it 
& 10.30 & 10.30 & 10.30
& 7.81 & 8.20 & 8.26
& 8.38 & 8.67 & 8.71 \\
gemma-3-12b-it 
& 9.70 & 9.70 & 10.30
& 7.06 & 7.61 & 8.49
& 7.62 & 8.04 & 8.27 \\
Llama-3.1-8B-Instruct 
& 6.06 & 6.06 & 6.06
& 4.32 & 4.73 & 4.82
& 4.47 & 4.77 & 4.80 \\
Llama-3.2-3B-Instruct 
& 3.94 & 4.24 & 4.55
& 3.23 & 3.68 & 3.98
& 2.88 & 2.95 & 2.94 \\
Fin-o1-14B 
& 0.00 & 0.00 & 0.00
& 0.00 & 0.00 & 0.00
& 0.00 & 0.00 & 0.00 \\
Fin-R1 
& 2.12 & 2.42 & 2.73
& 2.02 & 2.32 & 2.63
& 1.92 & 1.97 & 2.00 \\
\hline
\end{tabular}%
}
\end{table}

From Table~\ref{tab:sm-all}, we observe that retrieval performance on FinSM remains uniformly low across models, with no consistent advantage from larger parameter scales or domain-specific financial pretraining. 
Even state-of-the-art general-purpose LLMs achieve Hit Rate@20 below 13\%, while several models fail. 
This suggests that FinSM performance is not primarily driven by model size or surface-level domain familiarity, but instead depends on the ability to identify semantically inappropriate accounting concepts under strict taxonomy-defined constraints.

Notably, increasing model capacity yields marginal improvements, indicating implicit representation learning is insufficient when hierarchical, taxonomy-defined semantics are not explicitly modeled. 
Financial models show similarly limited gains, highlighting that exposure to financial text does not guarantee accurate detection of mis-tagged concepts within regulated taxonomies.

The qualitative analysis reveals that retrieval errors arise from confusion over which reported concepts are semantically inappropriate under the given reporting context. 
For example, a filing incorrectly reports \textit{CashAndCashEquivalentsAtCarryingValue} in a context where only \textit{Cash} is permitted.
In such cases, models often fail to flag the mismatch, instead retrieving tags or omitting the erroneous concept entirely. 
Similarly, models may overlook the misuse of aggregate tags such as \textit{AssetsCurrent} when more granular components are required by the disclosure context. 
These errors indicate that models tend to rely on surface-level semantic similarity, rather than reasoning about contextual validity and taxonomy-defined reporting granularity.

\subsection{FinRE}
Table~\ref{tab:re} indicates that FinRE is substantially more challenging than standard relation classification, as it requires precise interpretation of hierarchical, compositional, and cross-document constraints rather than surface-level semantic cues.

\begin{table}[!t]
\setlength{\abovecaptionskip}{3pt}  
 \caption{The overall performance (\%) under the zero-shot settings on the FinRE task.}
 \label{tab:re}
 \centering
 \tiny
 \resizebox{.9\linewidth}{!}{%
 \begin{tabular}{l|cccc}
 \hline
 LLMs & Acc & Macro P & Macro R & Macro F1 \\ \hline
 GPT-4o & 91.82 & 90.15 & 90.03 & 90.09 \\
 DeepSeek-V3 & 82.73 & 87.81 & 80.19 & 80.68 \\
 Qwen3-235B-A22B-Instruct-2507 & 62.73 & 69.22 & 67.18 & 63.56 \\
 Llama-4-Scout-17B-16E-Instruct & 27.50 & 17.09 & 35.82 & 22.71 \\
 Qwen2.5-72B-Instruct & 67.50 & 72.14 & 71.50 & 68.27 \\
 Llama-3.3-70B-Instruct & 28.86 & 23.06 & 38.26 & 21.54 \\
 Qwen3-32B & 0.00 & 0.00 & 0.00 & 0.00 \\
 gemma-3-27b-it & 45.91 & 35.97 & 53.00 & 40.85 \\
 gemma-3-12b-it & 27.95 & 31.78 & 25.81 & 27.37 \\
 Llama-3.1-8B-Instruct & 30.45 & 15.25 & 26.58 & 19.34 \\
 Llama-3.2-3B-Instruct & 17.73 & 22.31 & 12.74 & 15.70 \\
 Fin-o1-14B & 0.00 & 0.00 & 0.00 & 0.00 \\
 Fin-R1 & 32.73 & 34.05 & 21.00 & 25.97 \\
 \hline
 \end{tabular}%
 }
\end{table}

Among all evaluated models, GPT-4o exhibits the strongest and most stable performance, achieving high accuracy and balanced macro-level precision, recall, and F1.
This suggests that frontier models are capable of jointly reasoning over multiple relationship types when sufficient representational and reasoning capacity is available.
DeepSeek-V3 follows as the strongest open-source model, but with a noticeable performance gap, indicating reduced robustness across relation categories.

In contrast, most open-source models struggle to generalize to this setting.
Model scale alone does not lead to consistent improvements: several large models underperform smaller counterparts, and the Llama family exhibits uniformly weak results across scales.
Notably, domain-specific financial LLMs do not demonstrate a clear advantage, with one model failing entirely and another achieving only limited accuracy.
These observations suggest that relationship error classification in FinRE cannot be addressed by domain pretraining or parameter scaling alone, but instead requires explicit alignment with taxonomy-defined structural semantics.

\begin{figure}[!h]
    \centering
    \setlength{\abovecaptionskip}{2pt}
    \setlength{\belowcaptionskip}{0pt}
    \includegraphics[width=\linewidth]{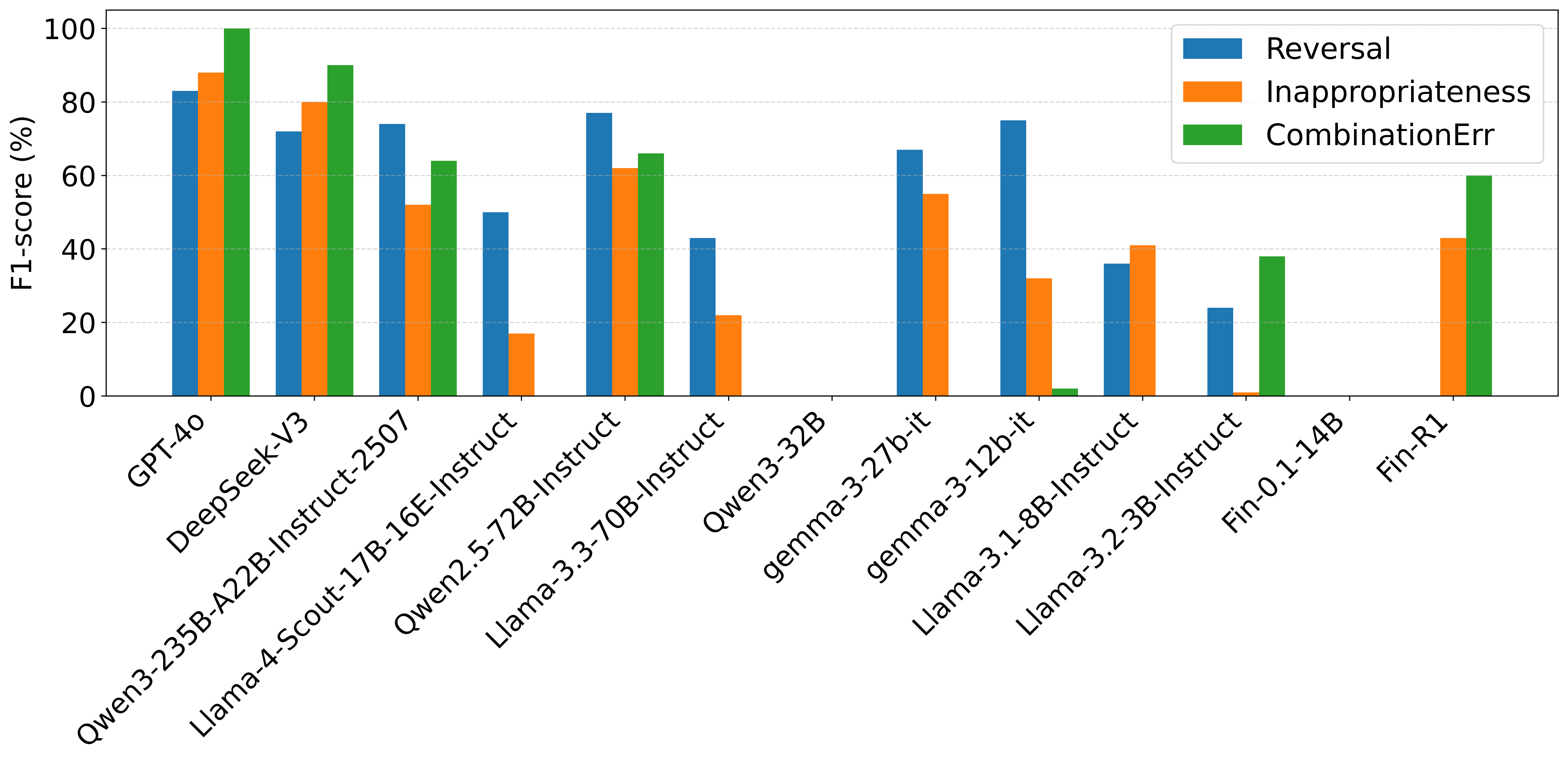}
    \caption{The F1-score (\%) for individual relation type under the zero-shot settings on the FinRE task.}
    \label{fig:re-ind}
\end{figure}


Finer-grained analysis by relation type (Figure~\ref{fig:re-ind}) reveals substantial difficulty variation.
While \textit{Reversal} and \textit{Inappropriateness} relations are handled reasonably well by top models, \textit{CombinationErr} emerges as the most challenging category.
Most models collapse to near-zero F1 on this relation type, indicating difficulty in validating axis–member combinations that require reasoning across multiple interrelated linkbases.
Even competitive models exhibit sharp performance drops, highlighting the complexity of enforcing structural consistency beyond pairwise relationships.


\subsection{FinMR}

\begin{figure}[t]
  \centering
  \begin{subfigure}[t]{\linewidth}
    \centering
    \includegraphics[width=\linewidth]{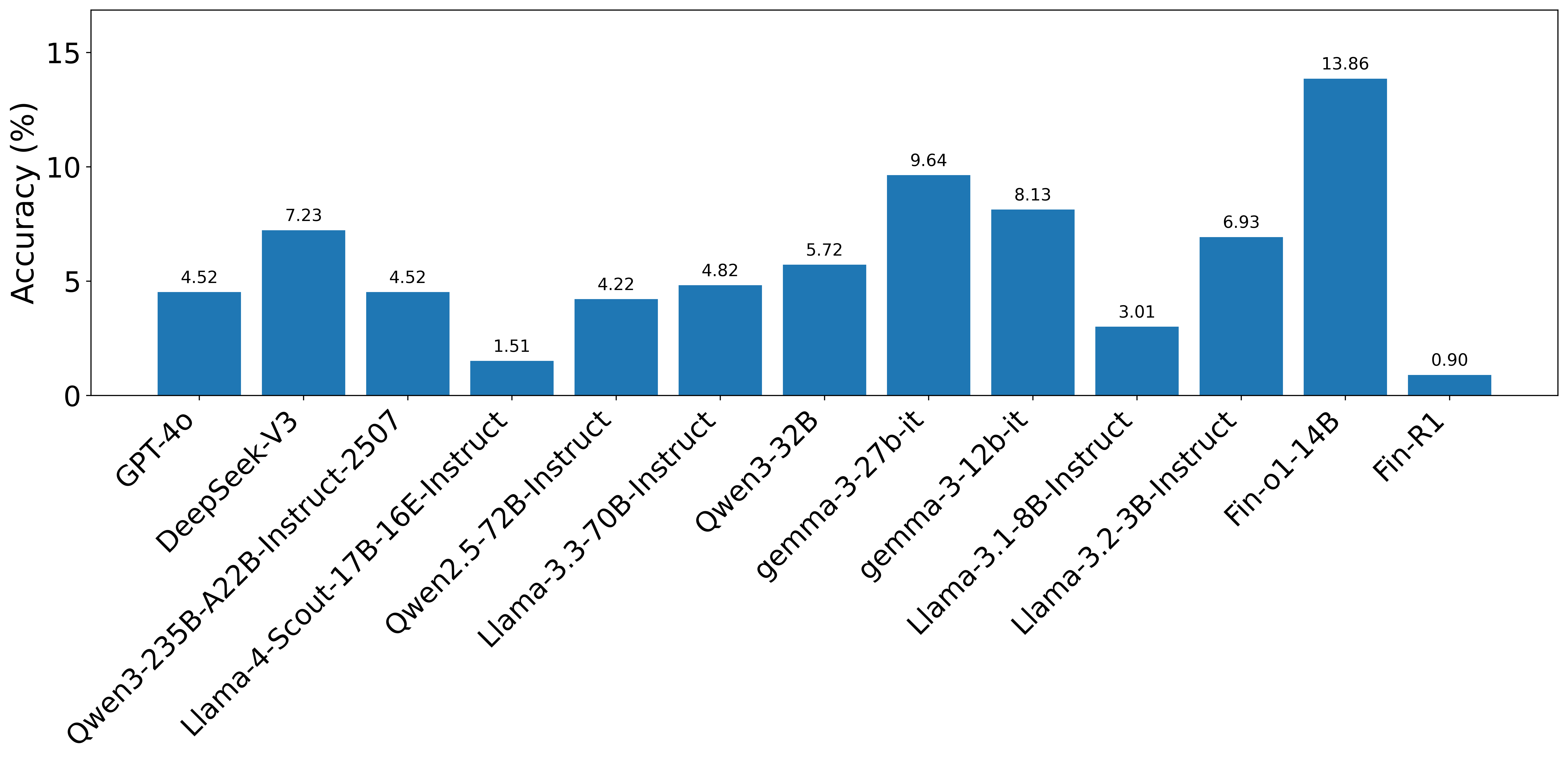}
    \caption{Overall accuracy under zero-shot settings.}
    \label{fig:finmr_acc}
  \end{subfigure}
  \hfill
  \begin{subfigure}[t]{\linewidth}
    \centering
    \includegraphics[width=\linewidth]{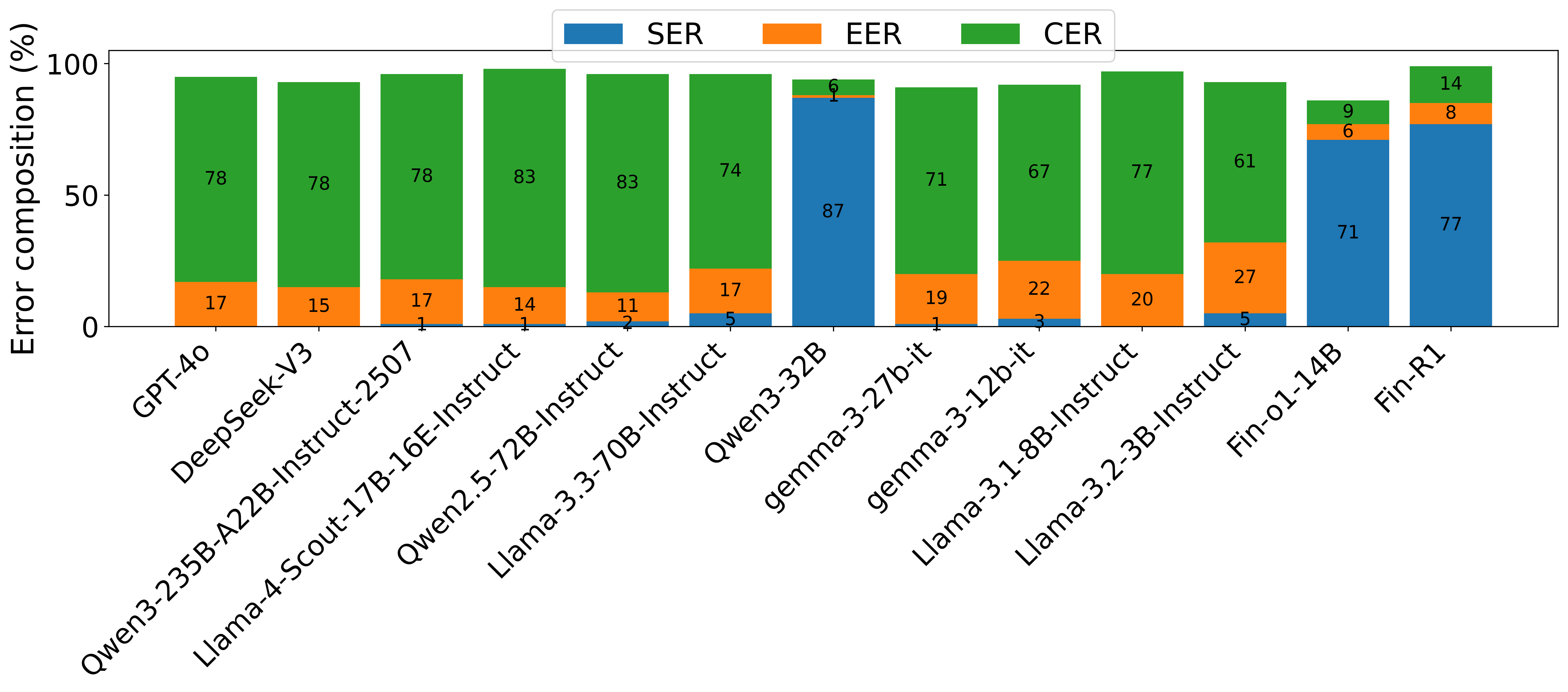}
    \caption{Error-rate composition, including SER, ERR, and CER.}
    \label{fig:finmr_error_rate}
  \end{subfigure}
  \caption{Zero-shot performance on the FinMR task.}
  \label{fig:finmr}
\end{figure}

Figure~\ref{fig:finmr} summarizes the zero-shot performance of representative large language models on the FinMR task. Overall accuracy remains consistently low, with the best-performing model, Fin-o1-14B, achieving only 13.86\% accuracy. 
This underscores the intrinsic difficulty of FinMR, which requires jointly reasoning over hierarchical calculation structures, cross-fact dependencies, and arithmetic constraints defined in XBRL filings, rather than answering isolated numerical questions.

The error composition analysis in Figure~\ref{fig:finmr_error_rate} provides a more fine-grained diagnostic view of model failures. 
Calculation errors (CER) dominate across most models, typically accounting for 70–83\% of failures. 
This indicates that while models may identify relevant financial facts, they frequently fail to execute or validate multi-step arithmetic relations specified by taxonomy-grounded calculation rules. 
Extraction errors (EER) are comparatively moderate (approximately 10–27\%), suggesting that locating candidate values is often feasible, but transforming them into correct numerical inferences remains a major bottleneck.

Structural errors (SER) further expose limitations in producing structure-compliant outputs. 
Models such as Qwen3-32B and Fin-R1 exhibit extremely high SER values, indicating frequent failures to generate outputs that conform to the required structured format. 
In contrast, Fin-o1-14B shows a markedly different error profile: despite achieving the highest overall accuracy and maintaining a low CER (9\%), it still suffers from a high SER (71\%). This pattern suggests that even when numerical reasoning is reliable, failures in adhering to the expected output format can prevent correct answers from being properly expressed and evaluated.

\section{Conclusion}

In this work, we present \textsc{\finauditing}, the first benchmark designed to evaluate LLMs on structured, hierarchical, and taxonomy-driven financial reasoning across multi-document filings. 
Built from real-world XBRL data, \textsc{\finauditing} defines three complementary tasks, FinSM, FinRE, and FinMR, which jointly assess structured semantic consistency, hierarchical relationship understanding, and multi-step mathematical reasoning. 
Our extensive zero-shot evaluation of state-of-the-art LLMs shows that even leading models, such as GPT-4o, DeepSeek-V3, and Fin-o1-14B, still struggle with cross-document reasoning, structured interpretation, and taxonomy alignment. 
These findings highlight fundamental limitations in current LLMs for financial auditing and emphasize the importance of developing models with stronger structural grounding and explicit reasoning awareness. 
We release all datasets and evaluation code to support future research on trustworthy financial intelligence.

\bibliographystyle{ACM-Reference-Format}
\bibliography{sample-base}

\appendix

\section{LLM Judgment for Data Annotation}
\label{annotation_rpompt}
\subsection{Prompt for the FinSM task}

\begin{tcolorbox}[
    colback=lightgray!10,
    colframe=black,
    title=DQC\_0099 Example,
    title style={fontupper=\scriptsize},
    width=\linewidth,       
    boxrule=0.4pt,
    arc=2pt,
    top=1pt,
    bottom=1pt,
    left=1pt,
    right=1pt,
    boxsep=1.5pt,
    before skip=4pt, 
    after skip=4pt
]
\tiny
\begin{lstlisting}[breaklines=true, basicstyle=\ttfamily\tiny]
You are given a DQC validation message. Your task is to extract two specific pieces of information:
1. **Statement Name**: Extract the name of the financial statement mentioned in the message. If it appears in a phrase like "0000007 - Statement - Consolidated Statements of Operations", ignore any prefixes such as "- Statement -" or ID numbers. Only return the clean name of the statement (e.g., "Consolidated Statements of Operations").
2. **Erroneous us-gaap Tag(s)**: From all `us-gaap:` prefixed tags mentioned in the message, extract only those that are clearly described as having an error or violation. A tag should be included **only if** it satisfies both of the following:
   - It is explicitly stated to be **not included in any calculation relationship**.
   - The message states that it **will produce an error**, **violates a rule**, or **should have been reported differently**.
Do not include `us-gaap:` tags that are merely suggested, referenced as valid abstract parents, or shown as examples for correction.
3. **Reasoning**: For each extracted erroneous tag, provide a short explanation summarizing why it was classified as erroneous, based on the message.
Output your result as a structured JSON in the following format:
```json
{
  "statement_name": "...",
  "error_tags": [
    {
      "tag": "...",
      "reason": "..."
    }
  ]
}
```
\end{lstlisting}
\end{tcolorbox}

\begin{tcolorbox}[
    colback=lightgray!10,
    colframe=black,
    title=DQC\_0109 Example,
    title style={fontupper=\scriptsize},
    width=\linewidth,       
    boxrule=0.4pt,
    arc=2pt,
    top=1pt,
    bottom=1pt,
    left=1pt,
    right=1pt,
    boxsep=1.5pt,
    before skip=4pt, 
    after skip=4pt
]
\tiny
\begin{lstlisting}[breaklines=true, basicstyle=\ttfamily\tiny]
You are given a DQC validation message. Your task is to extract all Dimensions information from the message.
1. Extract all key-value pairs under Dimensions, where each dimension is formatted as us-gaap:XXX=YYY.
2. For each extracted dimension, provide a short explanation in the "reason" field based on the message context. If no explicit issue is mentioned about a specific dimension, you may leave the "reason" as an empty string.
Output your result as a structured JSON in the following format:
```json
{
  "error_tags": [
    {
      "dimension": "...",
      "reason": "..."
    }
  ]
}
```
\end{lstlisting}
\end{tcolorbox}

\begin{tcolorbox}[
    colback=lightgray!10,
    colframe=black,
    title=DQC\_0137 Example,
    title style={fontupper=\scriptsize},
    width=\linewidth,       
    boxrule=0.4pt,
    arc=2pt,
    top=1pt,
    bottom=1pt,
    left=1pt,
    right=1pt,
    boxsep=1.5pt,
    before skip=4pt, 
    after skip=4pt
]
\tiny
\begin{lstlisting}[breaklines=true, basicstyle=\ttfamily\tiny]
You are given a DQC validation message. Your task is to extract the following information:
1. Target Tag: Extract the primary us-gaap tag being referenced or discussed as erroneous in the message. Only include the tag explicitly stated as problematic or requiring correction.
2. Dimensions: Extract all dimension key-value pairs listed under "Dimensions" in the message. Each dimension should be in the format us-gaap:XXX=YYY.
3. For each extracted dimension, provide a short explanation in the "reason" field based on the message context. If the message does not explicitly mention an issue with the dimension, you may leave the "reason" field empty or give a general summary.
Output your result as a structured JSON in the following format:
```json
{
  "target_tag": "...",
  "error_tags": [
    {
      "dimension": "...",
      "reason": "..."
    }
  ]
}
```
\end{lstlisting}
\end{tcolorbox}

\subsection{Prompt for the FinRE task}

\begin{tcolorbox}[
    colback=lightgray!10,
    colframe=black,
    title=DQC\_0081 Example,
    title style={fontupper=\scriptsize},
    width=\linewidth,       
    boxrule=0.4pt,
    arc=2pt,
    top=1pt,
    bottom=1pt,
    left=1pt,
    right=1pt,
    boxsep=1.5pt,
    before skip=4pt, 
    after skip=4pt
]
\tiny
\begin{lstlisting}[breaklines=true, basicstyle=\ttfamily\tiny]
You are given a DQC validation message. Your task is to extract two specific pieces of information:
1. **Statement Name**: Extract the name of the financial statement mentioned in the message. If it appears in a phrase like "00000002 - Statement - BALANCE SHEET", ignore any prefixes such as ID numbers or "- Statement -". Only return the clean name of the statement (e.g., "BALANCE SHEET").
2. **Erroneous us-gaap Tag Relationships**: From all `us-gaap:` tags mentioned in the message, identify pairs of tags where the relationship between them is described as incorrect or inconsistent. A tag pair should be extracted **only if** it satisfies both of the following conditions:
   - The two tags are stated to be in a specific structural relationship in the filer's taxonomy (e.g., sibling, parent-child).
   - The message clearly indicates that this relationship contradicts the definition in the official US-GAAP taxonomy, or requires reviewer attention for potential misclassification.
3. **Reasoning**: For each extracted tag pair, provide a short explanation summarizing the inconsistency or incorrect relationship between them as described in the message.
Output your result as a structured JSON in the following format:
```json
{
  "statement_name": "...",
  "error_tags": [
    {
      "tag1": "...",
      "tag2": "...",
      "reason": "..."
    }
  ]
}
```
\end{lstlisting}
\end{tcolorbox}

\begin{tcolorbox}[
    colback=lightgray!10,
    colframe=black,
    title=DQC\_0001 Example,
    title style={fontupper=\scriptsize},
    width=\linewidth,       
    boxrule=0.4pt,
    arc=2pt,
    top=1pt,
    bottom=1pt,
    left=1pt,
    right=1pt,
    boxsep=1.5pt,
    before skip=4pt, 
    after skip=4pt
]
\tiny
\begin{lstlisting}[breaklines=true, basicstyle=\ttfamily\tiny]
You are an expert in XBRL taxonomies, tasked with parsing DQC validation messages for relationship errors (Rule DQC.US.0001), specifically an axis with an inappropriate member.
1. **Main Concept**: Extract the primary concept that is being dimensionally qualified. This is usually the first concept mentioned in the message.
2. **Dimension Pair**: From the 'Dimensions' field in the message, extract the full axis and member combination string **exactly as it appears**, including all prefixes.
3. **Reasoning**: Provide a short explanation that the member is unallowable for the specified axis, as described in the message.
Output your result as a single, structured JSON object in the following format:
```json
{
  "main_concept": "UnrealizedGainLossOnCashFlowHedgingInstruments",
  "dimension_pair": "us-gaap:FairValueByFairValueHierarchyLevelAxis=us-gaap:FairValueMeasurementsRecurringMember",
  "reason": "The message indicates the member is unallowable for the specified axis."
}
```
\end{lstlisting}
\end{tcolorbox}

\begin{tcolorbox}[
    colback=lightgray!10,
    colframe=black,
    title=DQC\_0145 Example,
    title style={fontupper=\scriptsize},
    width=\linewidth,       
    boxrule=0.4pt,
    arc=2pt,
    top=1pt,
    bottom=1pt,
    left=1pt,
    right=1pt,
    boxsep=1.5pt,
    before skip=4pt, 
    after skip=4pt
]
\tiny
\begin{lstlisting}[breaklines=true, basicstyle=\ttfamily\tiny]
You are an expert in XBRL taxonomies, tasked with parsing DQC validation messages for relationship errors (Rule DQC.US.0145), specifically 'Inappropriate Cash Flow Presentation'.
1. **Head Concept**: Extract the primary concept that is inappropriately presented. This is usually the first concept mentioned in the message.
2. **Tail Concept**: Extract the presentation concept that the head concept is incorrectly a descendant of.
3. **Reasoning**: Provide a short explanation that the head concept should not be presented as a component of the tail concept.
Output your result as a single, structured JSON object in the following format:
```json
{
  "head_concept": "element 1",
  "tail_concept": "element 2",
  "reason": "The message indicates the head concept is inappropriately presented as a descendant of the tail concept and should be outside this group."
}
```
\end{lstlisting}
\end{tcolorbox}

\subsection{prompt for the FinMR task}

\begin{tcolorbox}[
    colback=lightgray!10,
    colframe=black,
    title=DQC\_0015 Example,
    title style={fontupper=\scriptsize},
    width=\linewidth,       
    boxrule=0.4pt,
    arc=2pt,
    top=1pt,
    bottom=1pt,
    left=1pt,
    right=1pt,
    boxsep=1.5pt,
    before skip=4pt, 
    after skip=4pt
]
\tiny
\begin{lstlisting}[breaklines=true, basicstyle=\ttfamily\tiny]
You are given a DQC validation message. Your task is to extract the following key information:
1. **Erroneous Tag Information:**:
   Identify the primary us-gaap tag in the message that is reported with an incorrect value due to sign (positive/negative) error. For this tag, extract: 
   - `tag`: The us-gaap element tag with the error.
   - `period`: The reporting period for this tag, as shown in the "Period" field.
   - `reported_value`: The actual value reported in the filing, as stated in the message or "Value" field.
   - `correct_value`: The correct value, which should be the absolute value of the reported value (i.e., reported_value without the negative sign), as implied or explicitly required by the message.
   - `reason`: Briefly explain why the reported value is considered incorrect, based on the message content (for example, "This element should not have a negative value. The amount should be input as a positive value and, if necessary, a negated label should be provided.").  
Return your result as structured JSON in the following format:
```json
{
  "error_tags": 
    {
      "tag": "...",
      "period": "...",
      "reported_value": "...",
      "correct_value": "...",
      "reason": "..."
    }
}
```
\end{lstlisting}
\end{tcolorbox}

\begin{tcolorbox}[
    colback=lightgray!10,
    colframe=black,
    title=DQC\_0117 Example,
    title style={fontupper=\scriptsize},
    width=\linewidth,       
    boxrule=0.4pt,
    arc=2pt,
    top=1pt,
    bottom=1pt,
    left=1pt,
    right=1pt,
    boxsep=1.5pt,
    before skip=4pt, 
    after skip=4pt
]
\tiny
\begin{lstlisting}[breaklines=true, basicstyle=\ttfamily\tiny]
You are an expert financial analyst parsing DQC validation messages for mathematical reasoning errors (Rule DQC.US.0117). Your task is to extract all required components from the message.
1. **Statement Name**: Extract the clean name of the financial statement (e.g., "Condensed Consolidated Income Statements").
2. **Input Data**: From the properties list at the end of the message, extract the following:
    * `target_concept`: The full name of the primary `us-gaap:` concept being evaluated.
    * `period`: The reporting period for the fact (e.g., "2020-10-01 to 2020-12-31").
3. **Output Data**: From the main body of the message, extract the following values:
    * `extracted_value`: The value of the concept as reported in the filing (e.g., "255,500,000").
    * `calculated_value`: The correct value based on the dimensional breakdown (e.g., "142,400,000").
    * `is_correct`: This should always be "No" for these error messages.
Output your result as a single, structured JSON object in the following format:
```json
{
  "statement_name": "...",
  "input": {
    "target_concept": "us-gaap:RevenueFromContractWithCustomerIncludingAssessedTax",
    "period": "..."
  },
  "output": {
    "extracted_value": "...",
    "calculated_value": "...",
    "is_correct": "No"
  }
}
```
\end{lstlisting}
\end{tcolorbox}

\begin{tcolorbox}[
    colback=lightgray!10,
    colframe=black,
    title=DQC\_0126 Example,
    title style={fontupper=\scriptsize},
    width=\linewidth,       
    boxrule=0.4pt,
    arc=2pt,
    top=1pt,
    bottom=1pt,
    left=1pt,
    right=1pt,
    boxsep=1.5pt,
    before skip=4pt, 
    after skip=4pt
]
\tiny
\begin{lstlisting}[breaklines=true, basicstyle=\ttfamily\tiny]
You are given a DQC validation message. Your task is to extract the following key information:
1. **Statement Name**: Extract the name of the financial statement mentioned in the message. If it appears in a phrase like "000004 - Statement - CONSOLIDATED STATEMENTS OF OPERATIONS (Unaudited)", ignore any numeric prefixes and the phrase "- Statement -". Only return the clean name of the statement, such as "CONSOLIDATED STATEMENTS OF OPERATIONS (Unaudited)".
2. **Erroneous Total Element Tag Information**:
   Extract detailed information about the total element (`us-gaap:` tag) that is reported incorrectly. Specifically, you should extract:
   - `tag`: The total element tag, as indicated by the **"Total Element"** field in the message.
   - `period`: The reporting period associated with this tag, as indicated by the **"Total period"** field.
   - `reported_value`: The value that was reported in the filing, taken from the **"Total Value"** field.
   - `correct_value`: The expected value, i.e., the **sum of the child components** as defined in the calculation linkbase.
     - This is usually explicitly stated in a sentence like: *"The sum of these child components is..."*
     - You may cross-validate this by summing the listed child component values in the message if provided.
   - `reason`: Provide a concise explanation of the error - why the reported value is considered incorrect - based directly on the message content.
Only include this tag if all of the following are true:
- The message identifies a total element with incorrect calculation
- Both reported and correct values are stated or inferable
- The error occurs in a specified reporting period
Return your result as structured JSON in the following format:
```json
{
  "statement_name": "...",
  "error_tags": 
    {
      "tag": "...",
      "period": "...",
      "reported_value": "...",
      "correct_value": "...",
      "reason": "..."
    }
}
```
\end{lstlisting}
\end{tcolorbox}

\section{US-GAAP Taxonomy Chunking}
\label{taxonomy_chunking}

To enable structured retrieval and reasoning, we convert the official US-GAAP taxonomy spreadsheets into \textbf{concept-centric chunks}. Each GAAP concept is represented by a \textit{core chunk} capturing its intrinsic attributes and a set of \textit{relation chunks} describing its semantic and structural links to other concepts. The conversion operates directly on the official Excel workbook through systematic normalization and extraction.

\paragraph{Sheet Normalization.}
All sheets are loaded from the taxonomy workbook, column headers are standardized, and relevant sheets (\textsf{Concept}, \textsf{Presentation}, \textsf{Calculation}, \textsf{Definition}, \textsf{Reference}, \textsf{Enumeration}) are identified by keyword matching. Unnamed or irrelevant reference sheets are excluded, and each extracted element is annotated with provenance information such as file name, sheet name, and row number.

\paragraph{Core Chunk Extraction.}
From the \textsf{Concepts} sheet, each concept’s identifier, label, type, balance, period type, abstract flag, documentation, and deprecation status are extracted and rendered into a canonical text form (\textsf{concept::core}) suitable for retrieval or embedding.

\paragraph{Relation Chunk Extraction.}
For each concept, all relation-bearing sheets are parsed to construct chunks representing presentation hierarchies, calculation formulas, definition arcs, reference citations, and enumerations. The parser handles heterogeneous layouts, missing columns, and varying role conventions. Each chunk is recorded with its role, arcrole, and provenance metadata under the identifier \textsf{concept::relations:{pres|calc|def|ref|enum}}, with integrity checks ensuring that all relation targets correspond to valid core concepts.

\paragraph{Outputs.}
All chunks are written into \textsf{chunks\_core.jsonl} and \textsf{chunks\_relations.jsonl}, and summary statistics are stored in \textsf{meta.json}. This representation preserves the hierarchical and relational structure of the US-GAAP taxonomy while making each concept independently retrievable.

To enable structured retrieval and reasoning, we convert the official US-GAAP taxonomy spreadsheets into \textbf{concept-centric chunks}. Each GAAP concept is represented by two complementary components: a \textit{core chunk} that captures intrinsic attributes and a collection of \textit{relation chunks} that encode its semantic and structural links to other concepts. The conversion process operates directly on the official Excel workbook, performing systematic normalization and extraction as follows.

\paragraph{Sheet Normalization and Identification.} 
We first load all sheets within the taxonomy workbook, standardize their column headers by lowercasing and trimming whitespace, and identify relevant sheets by keyword matching (\textsf{Concept}, \textsf{Presentation}, \textsf{Calculation}, \textsf{Definition}, \textsf{Reference}, and \textsf{Enumeration}). Suspicious ``all-unnamed'' reference sheets are ignored. The taxonomy version number is automatically inferred from the filename. Each extracted element is annotated with its provenance, including the file name, sheet name, and row number.

\paragraph{Core Chunk Extraction.}
From the \textsf{Concepts} sheet, we extract for each concept its identifier (\textsf{prefix:name}), label, type, balance, period type, abstract flag, documentation, and deprecation status. These attributes are rendered into a canonical, human-readable \textsf{chunk\_text} and stored with a stable identifier in the form \textsf{concept::core}. The core chunk provides a self-contained textual summary suitable for direct retrieval or embedding.

\paragraph{Relation Chunk Extraction.}
For every concept, we parse all relation-bearing sheets to construct one chunk per relation family:
\begin{itemize}
    \item \textbf{Presentation:} hierarchical parent--child paths with associated roles and preferred labels.
    \item \textbf{Calculation:} quantitative relationships including parent/child direction, weights, and roles.
    \item \textbf{Definition:} arcs between dimensions, domains, and hypercubes, including arcrole, source, and target.
    \item \textbf{Reference:} external standard citations consisting of source, section, and note, filtered to the relevant concept.
    \item \textbf{Enumeration:} extensible lists aggregated by domain and linkrole, with de-duplicated members.
\end{itemize}

The extraction is designed to handle heterogeneous sheet layouts and missing columns: when explicit \textsf{from*}/\textsf{to*} fields are absent, the parser falls back to parent/child-style columns and merges multiple role or linkrole conventions. Each relation chunk records the roles and arcroles encountered, as well as complete provenance metadata, and uses the identifier \textsf{concept::relations:\{pres|calc|def|ref|enum\}} to store. Integrity validation ensures that all relation targets reference existing core concepts, and malformed rows are automatically skipped.

\paragraph{Outputs.}
All chunks are written into \textsf{chunks\_core.jsonl} and \textsf{chunks\_relations.jsonl}, while global statistics are summarized in \textsf{meta.json}. This chunked representation preserves the hierarchical and relational organization of the US-GAAP taxonomy while making each concept addressable as an atomic, retrievable unit, as shown in the following.

\begin{tcolorbox}[
  colback=lightgray!5,
  colframe=black,
  title=Example of a Core Chunk,
  title style={fontupper=\scriptsize},
  width=\linewidth,
  boxrule=0.4pt,
  arc=2pt,
  top=1pt,
  bottom=1pt,
  left=1pt,
  right=1pt,
  boxsep=1.5pt,
  before skip=4pt,
  after skip=4pt
]
\tiny
\begin{lstlisting}[breaklines=true, basicstyle=\ttfamily\tiny]
[Concept Core]
ID: us-gaap:Revenues
Label: Revenues
Type: monetaryItemType | Balance: credit | PeriodType: duration | Abstract: False
Status: active | DeprecatedLabel: None | DeprecatedDate: None
Documentation:
Amount of revenue recognized from goods sold or services rendered during the period.
Provenance: file=GAAP_Taxonomy_2024.xlsx | sheet=Concepts | row=248
\end{lstlisting}
\end{tcolorbox}

\begin{tcolorbox}[
  colback=lightgray!5,
  colframe=black,
  title=Example of a Relation Chunk,
  title style={fontupper=\scriptsize},
  width=\linewidth,
  boxrule=0.4pt,
  arc=2pt,
  top=1pt,
  bottom=1pt,
  left=1pt,
  right=1pt,
  boxsep=1.5pt,
  before skip=4pt,
  after skip=4pt
]
\tiny
\begin{lstlisting}[breaklines=true, basicstyle=\ttfamily\tiny]
[Concept Relations]
ID: us-gaap:Revenues
Presentation:
- Role: Statement - Income Statement
  Path: Revenues > CostOfRevenue > GrossProfit
Calculation:
- As Parent:
  child=us-gaap:CostOfRevenue | weight=-1.0 | role=Calculation Linkbase
- As Parent:
  child=us-gaap:GrossProfit | weight=+1.0 | role=Calculation Linkbase
Definition (Dimensions/Domain/Hypercube):
- domain-member: from=RevenueRecognitionPolicyTextBlock -> to=RevenueCategoryAxis
References:
- FASB ASC 606 | Section 25 | Revenue Recognition
Provenance: sheets include Presentation-like and Calculation-like sheets
\end{lstlisting}
\end{tcolorbox}

\section{Evaluation Metrics}
\label{eval_metrics_appendix}

\subsection{The metrics for FinSM}

For the FinSM task, we evaluated retrieval performance using Hit Rate ($HR@k$), Recall ($R@k$), and Macro-F1 ($MF1@k$), where $k=\{1,5,10,20\}$. \textit{Hit Rate} evaluates whether the LLM retrieves at least one relevant element within the top-$k$ results. It reflects the proportion of queries for which the model can successfully hit a relevant item, without considering how many are retrieved. \textit{Recall} measures the fraction of all relevant elements that are successfully retrieved within the top-$k$ predictions. It captures how completely the model covers the relevant set under a cutoff of $k$. \textit{Macro-F1} balances precision and recall at the query level by computing the F1 score for each query using its top-$k$ predictions and then averaging across all queries. It reflects both the accuracy (precision) and completeness (recall) of retrieval, giving equal weight to each query. 
\begin{equation}
\scriptsize
    \mathrm{HR}@k = \frac{1}{N}\sum_{i=1}^N \mathbf{1}\big(E_i^{(k)} \cap G_i \neq \emptyset\big)
\end{equation}

\begin{equation}
\scriptsize
    \mathrm{R}@k = \frac{1}{N} \sum_{i=1}^N \frac{|E_i^{(k)} \cap G_i|}{|G_i|}
\end{equation}

\begin{equation}
\scriptsize
    \mathrm{Macro\text{-}F1}@k = \frac{1}{N'} \sum_{i=1}^{N'} 
    \frac{2 \times P_i^{(k)} \times R_i^{(k)}}{P_i^{(k)} + R_i^{(k)}}
\end{equation}
where
{\scriptsize
\begin{align*}
    P_i^{(k)} &= 
        \begin{cases}
            \dfrac{|E_i^{(k)} \cap G_i|}{|E_i^{(k)}|}, & |E_i^{(k)}| > 0 \\
            0, & \text{otherwise}
        \end{cases}, \quad
    R_i^{(k)} =
        \begin{cases}
            \dfrac{|E_i^{(k)} \cap G_i|}{|G_i|}, & |G_i| > 0 \\
            0, & \text{otherwise}
        \end{cases}
\end{align*}
}

and $N$ is the total number of queries, and $N'$ is the number of queries with non-empty ground truth sets. $E_i^{(k)}$ denotes the top-$k$ predicted elements for the $i$-th query. $G_i$ denotes the set of ground truth elements for the $i$-th query. $|E_i^{(k)} \cap G_i|$ is the number of correctly predicted elements among the top-$k$ results.  $k$ is the cutoff rank indicating the number of top predictions considered. $\mathbf{1}(\cdot)$ is the indicator function, returning 1 if the condition holds and 0 otherwise.

\subsection{The metrics for FinRE}

For the FinRE task, we assessed the performance of various LLMs using Accuracy (Acc), Precision (Macro P), Recall (Macro R), and F1-score (Macro F1). Accuracy measures the overall proportion of correctly classified relations among all predictions. Precision evaluates the proportion of correctly predicted relations within each class, averaged across the three relation categories (Macro P). Recall quantifies the proportion of true relations that were successfully identified for each class, also averaged across classes (Macro R). F1-score provides a balanced measure of Precision and Recall by computing their harmonic mean per class and then averaging across the three classes (Macro F1).

\begin{equation}
\scriptsize
    \mathrm{Acc} = \frac{1}{N} \sum_{i=1}^{N} \mathbf{1}(\hat{y}_i = y_i)
\end{equation}

\begin{equation}
\scriptsize
    \mathrm{Macro\text{-}P} = \frac{1}{C} \sum_{c=1}^{C} 
    \frac{TP_c}{TP_c + FP_c}
\end{equation}

\begin{equation}
\scriptsize
    \mathrm{Macro\text{-}R} = \frac{1}{C} \sum_{c=1}^{C} 
    \frac{TP_c}{TP_c + FN_c}
\end{equation}

\begin{equation}
\scriptsize
    \mathrm{Macro\text{-}F1} = \frac{1}{C} \sum_{c=1}^{C} 
    \frac{2 \times P_c \times R_c}{P_c + R_c}
\end{equation}
where $N$ is the total number of instances. $C$ is the number of relation categories (here $C=3$). $y_i$ and $\hat{y}_i$ denote the ground truth and predicted relation labels for the $i$-th instance. $TP_c$, $FP_c$, and $FN_c$ denote the number of true positives, false positives, and false negatives for class $c$, respectively. $P_c$ and $R_c$ denote the precision and recall for class $c$. $\mathbf{1}(\cdot)$ is the indicator function that equals 1 when the prediction is correct and 0 otherwise.

\subsection{The metrics for FinMR}
For the \textbf{FinMR} task, we evaluated the mathematical reasoning capabilities of various LLMs using an \textit{LLM-as-a-judge} framework to assess the overall \textit{Accuracy} of generated answers. This metric reflects whether each model produces the correct numerical result under the specified auditing rule. In addition to overall correctness, we introduced three fine-grained error indicators, namely the \textit{Structural Error Rate} (\textit{SER}), \textit{Extraction Error Rate} (\textit{EER}), and \textit{Calculation Error Rate} (\textit{CER}), each corresponding to a specific stage of reasoning validation.

\textit{SER} measures the proportion of outputs with invalid structure, where the model fails to produce the required JSON format containing both the extracted and calculated values. \textit{EER} quantifies errors in identifying or reproducing the correct extracted value, meaning that the predicted numerical entity does not match the true extracted value in mathematical meaning. \textit{CER} captures computational mistakes, where the predicted calculated value deviates from the ground truth even when the structure and extracted value are both correct. Together, these indicators provide a hierarchical view of reasoning quality, reflecting whether an error arises from structural formatting, factual extraction, or numerical computation, thereby offering fine-grained insights into model performance on complex financial mathematical reasoning tasks. The specific prompt for judgment is shown below.

\begin{tcolorbox}[
  colback=lightgray!5,
  colframe=black,
  title=LLM-as-a-judge prompt for FinMR Task,
  title style={fontupper=\scriptsize},
  width=\linewidth,
  boxrule=0.4pt,
  arc=2pt,
  top=1pt,
  bottom=1pt,
  left=1pt,
  right=1pt,
  boxsep=1.5pt,
  before skip=4pt,
  after skip=4pt
]
\tiny
\begin{lstlisting}[breaklines=true, basicstyle=\ttfamily\tiny]
Instruction: You are an evaluator. Your task is to judge whether the model's output pred_answer is correct compared to the given true_answer. 
Follow the rules strictly:
Step 1 (Structure Check):
    Verify whether pred_answer has the same structure as true_answer. The required structure is a JSON object with exactly two keys:
        {{"extracted_value": <value>, "calculated_value": <value>}}
    Minor formatting differences (e.g., line breaks, indentation, whitespace) are acceptable.
    If the structure is invalid, output the label: S
    If valid, continue to Step 2
Step 2 (Extracted Value Check):
    Compare true_answer["extracted_value"] and pred_answer["extracted_value"] by their mathematical meaning, not their string form. For example, "-1,284" and "-1284" are considered equal.
    If they are not equal in numeric meaning, output the label: E
    If equal, continue to Step 3
Step 3 (Calculated Value Check):
    Compare true_answer["calculated_value"] and pred_answer["calculated_value"] strictly in numeric meaning. They must be exactly equal (zero tolerance).
    If they are not equal, output the label: C
    If equal, then everything is correct
Final Decision:
    If all three checks pass, output the label: A
Output only one label: S, E, C, or A. Do not explain your reasoning.
\end{lstlisting}
\end{tcolorbox}

Given a total of $N$ evaluated instances, let $N_A$, $N_S$, $N_E$, and $N_C$ denote the numbers of instances labeled as \textit{Accurate (A)}, \textit{Structural Error (S)}, \textit{Extraction Error (E)}, and \textit{Calculation Error (C)}, respectively. 
Let $N_{\text{valid}}$ represent the number of successfully parsed instances that received a valid label. 
We define the following metrics:

\begin{equation}
\scriptsize
\text{Accuracy (ACC)} = \frac{N_A}{N_{\text{valid}}}
\end{equation}

\begin{equation}
\scriptsize
\text{Structural Error Rate (SER)} = \frac{N_S}{N_{\text{valid}}}
\end{equation}

\begin{equation}
\scriptsize
\text{Extraction Error Rate (EER)} = \frac{N_E}{N_{\text{valid}}}
\end{equation}

\begin{equation}
\scriptsize
\text{Calculation Error Rate (CER)} = \frac{N_C}{N_{\text{valid}}}
\end{equation}
where each instance is judged by the LLM-as-a-judge framework to output exactly one label from $\{A, S, E, C\}$. ACC, SER, EER, and CER, respectively, quantify the proportions of correct, structural, extraction, and calculation outcomes among validly parsed cases.

\section{Evaluation Models}
\label{eval_models_appendix}

Table~\ref{tab:model-table} summarizes all models evaluated in this study, categorized by openness, domain specialization, and architectural foundation. The evaluation covers a wide range of both closed- and open-source large language models (LLMs) as well as domain-specific and pretrained baselines.

\begin{table}[!h]
\scriptsize
\caption{Model categories and corresponding repositories.}
\label{tab:model-table}
\centering
\resizebox{\linewidth}{!}{%
\begin{tabular}{lll}
\toprule
\textbf{Model} & \textbf{Size} & \textbf{Repository} \\
\midrule
\textbf{Closed-source LLMs} \\
GPT-4o & -- & gpt-4o-2024-08-06 \\
\midrule
\textbf{Open-source LLMs} \\
DeepSeek-V3 & 685B & deepseek-chat \\
Llama-4-Scout-17B-16E-Instruct & 109B & meta-llama/Llama-4-Scout-17B-16E-Instruct \\
Qwen3-235B-A22B-Instruct-2507 & 235B & Qwen/Qwen3-235B-A22B-Instruct-2507 \\
Llama-3.3-70B-Instruct & 70B & meta-llama/Llama-3.3-70B-Instruct \\
Qwen2.5-72B-Instruct & 72B & Qwen/Qwen2.5-72B-Instruct \\
Qwen3-32B & 32B & Qwen/Qwen3-32B \\
gemma-3-27b-it & 27B & google/gemma-3-27b-it \\
gemma-3-12b-it & 12B & google/gemma-3-12b-it \\
Llama-3.1-8B-Instruct & 8B & meta-llama/Llama-3.1-8B-Instruct \\
Llama-3.2-3B-Instruct & 3B & meta-llama/Llama-3.2-3B-Instruct \\
\midrule
\textbf{Financial-specific LLMs} \\
Fin-o1-14B & 14B & TheFinAI/Fin-o1-14B \\
Fin-R1 & 7B & SUFE-AIFLM-Lab/Fin-R1 \\
\bottomrule
\end{tabular}%
}
\end{table}

\begin{itemize}
    \item \textbf{Closed-source LLMs:} We include GPT-4o~\cite{hurst2024gpt}, accessed via OpenAI’s API, as a representative of state-of-the-art proprietary models. Although the architecture and model size remain undisclosed, GPT-4o serves as a strong upper-bound reference in our benchmark.
    
    \item \textbf{Open-source LLMs:} This category encompasses a diverse set of publicly available, instruction-tuned models, including DeepSeek-V3~\cite{liu2024deepseek}, Qwen3-235B-A22B-Instruct-2507~\cite{qwen3technicalreport}, Qwen2.5-72B-Instruct~\cite{qwen2}, and Qwen3-32B~\cite{qwen3technicalreport}, offering a wide scale range. We also include several Llama models from Meta: Llama-4-Scout-17B-16E-Instruct~\cite{meta2025llama}, Llama-3.3-70B-Instruct~\cite{grattafiori2024llama}, Llama-3.1-8B-Instruct~\cite{grattafiori2024llama}, and Llama-3.2-3B-Instruct~\cite{grattafiori2024llama}, as well as Google's Gemma models~\cite{gemma_2024} (gemma-3-27b-it and gemma-3-12b-it). These models collectively enable cross-architecture and scaling comparisons across modern open-source systems.
    
    \item \textbf{Financial-specific LLMs:} We include Fin-o1-14B~\cite{qian2025fino1} and Fin-R1~\cite{liu2025finr1largelanguagemodel}, both trained on domain-specific financial corpora to capture specialized terminology and reasoning patterns in financial reporting. These models allow us to assess the impact of domain adaptation on structured auditing and numerical reasoning tasks.
\end{itemize}

Together, these models form a comprehensive evaluation spectrum, spanning closed- and open-source systems, general-purpose and domain-specialized models, and encoder- versus decoder-based architectures, enabling an in-depth analysis of performance across all proposed benchmark tasks.

\section{Practical Deployment and Scalability Considerations}
\label{practice}

This section provides additional discussion on how FINAUDITING may be integrated into real-world auditing workflows and scaled under practical constraints. The goal is not to prescribe a specific deployment system, but to illustrate how the benchmark design aligns with existing auditing practices and supports modular instantiation.

\paragraph{Integration into Existing Auditing Pipelines}
In practice, financial auditing commonly relies on rule-based validation systems, such as XBRL consistency checks and Data Quality Committee (DQC) rules, followed by targeted human review. FINAUDITING is naturally positioned to operate downstream of such validators. Rather than processing entire filings, models evaluated under FINAUDITING are applied only to a subset of suspicious items already identified by deterministic checks. Within this setting, the three benchmark tasks correspond to distinct stages of post-validation analysis. FinSM supports taxonomy disambiguation by assisting auditors in identifying appropriate concepts when potential mis-tagging is detected. FinRE provides structured categorization of detected issues into interpretable error types, which can guide remediation and prioritization. FinMR focuses on targeted numerical verification by tracing calculation structures and comparing reported values with recomputed ones, and is intended for precision-critical cases. This workflow reflects common auditing practices in which automated components emphasize recall, while high-stakes decisions remain subject to expert review.

\paragraph{Scalability across Regulatory Environments}
Although instantiated using the US-GAAP taxonomy, the task formulation of FINAUDITING is not tied to any specific accounting standard. The benchmark abstracts auditing as taxonomy-grounded reasoning over instance documents, schema definitions, and linkbase relations. Extending FINAUDITING to other regulatory environments, such as IFRS or jurisdiction-specific reporting standards, primarily requires substituting the underlying taxonomy and rule sources while preserving the same task interfaces. This decoupling between task definitions and accounting standards enables scalable adaptation without redesigning the benchmark structure.

\paragraph{Lightweight Variants and Semi-automatic Deployment}
FINAUDITING also supports tiered or hybrid deployment strategies that balance performance with computational cost. Full-capacity large language models are not required at all auditing stages. For example, FinSM can be instantiated using embedding-based retrieval or compact instruction-tuned models to achieve high recall at low cost, effectively narrowing the candidate space for auditors. FinRE involves a small and fixed label set, making it suitable for fine-tuning lightweight classifiers. FinMR, which requires precise numerical reasoning and evidence tracing, can be selectively applied to a limited number of high-risk cases and handled by larger models or human-in-the-loop verification. Such semi-automatic configurations enable practical deployment by combining lightweight automation with selective use of more advanced reasoning components.

Overall, by aligning with existing auditing workflows and supporting model stratification, FINAUDITING provides a modular evaluation framework that reflects how machine-assisted financial auditing systems may be integrated and scaled under realistic operational constraints.

\end{document}